\documentclass[sigconf, nonacm]{acmart}
\AtBeginDocument{%
  }

\setcopyright{none}
\renewcommand\footnotetextcopyrightpermission[1]{}

\copyrightyear{2026}
\acmYear{2026}
\acmDOI{XXXXXXX.XXXXXXX}
\acmConference[ICAIF '26]{The 7th ACM International Conference on AI in Finance}{November 14--17,
  2026}{Milan, Italy}
\acmISBN{978-1-4503-XXXX-X/2026/11}

\usepackage{pifont}     % for \ding tick and cross
\usepackage{makecell}   
\usepackage{enumitem}
\usepackage{subcaption} 
\usepackage{amsmath}
\usepackage{tikz}
\usepackage{booktabs}

\usepackage[table]{xcolor}
\usepackage{multirow}

\newcommand{\ie}{\emph{i.e., }}
\newcommand{\eg}{\emph{e.g., }}

\usetikzlibrary{positioning, arrows.meta, fit, backgrounds, calc,
    decorations.pathreplacing, shadows.blur, shapes.geometric}
\pgfdeclarelayer{deepbg}
\pgfsetlayers{deepbg,background,main}
\providecommand{\footref}[1]{}

\makeatletter
\patchcmd{\@mkauthors@iii}{\lineskip=1pc}{\lineskip=9.5pt}{}{%
  \ClassWarningNoLine{acmart}{Author-row spacing patch did not apply}}
\makeatother

\begin{document}

%%
%% The "title" command has an optional parameter,
%% allowing the author to define a "short title" to be used in page headers.
\title{Don't Model the Price, Model the Investors: Price Forecasting through Agent Role Learning with On-Policy Distillation}

\author{Kelvin J.L. Koa}
\affiliation{
  \institution{National University of Singapore,
               Asian Institute of Digital Finance
               }
  \country{Singapore}
  }
\email{kelvin.koa@u.nus.edu}

\author{Ziling Liu}
% \authornote{These authors contributed equally to this research.}
\affiliation{
  \institution{National University of Singapore,
               Asian Institute of Digital Finance
               }
  \country{Singapore}
  }
\email{e1547130@u.nus.edu}

\author{Yuncong Liu}
\affiliation{
  \institution{National University of Singapore,
               Asian Institute of Digital Finance
               }
  \country{Singapore}
  }
\email{yuncongliu@u.nus.edu}

\author{Zhengxi Qian}
\affiliation{
  \institution{National University of Singapore,
               Asian Institute of Digital Finance
               }
  \country{Singapore}
  }
\email{zhengxi4763@gmail.com}

\author{Shengqiong Wu}
\affiliation{%
  \institution{University of Oxford}
  \city{Oxford}
  \country{United Kingdom}
}
\email{shengqionggwu@gmail.com}

\author{Michael Wooldridge}
\affiliation{
  \institution{University of Oxford}
  \city{Oxford}
  \country{United Kingdom}
  }
\email{michael.wooldridge@cs.ox.ac.uk}

\author{Ke-Wei Huang}
\affiliation{%
  \institution{National University of Singapore,
               Asian Institute of Digital Finance
               }
  \country{Singapore}
  }
\email{dishkw@nus.edu.sg}

%%
%% Concise author list for page headers.
% \renewcommand{\shortauthors}{Koa et al.}

%%
%% By default, the full list of authors will be used in the page
%% headers. Often, this list is too long, and will overlap
%% other information printed in the page headers. This command allows
%% the author to define a more concise list
%% of authors' names for this purpose.

%%
%% The abstract is a short summary of the work to be presented in the
%% article.

\begin{abstract}
To date, there is limited theoretical justification to show that there exists any reliable way to reason over market information to forecast asset prices. In this work, we propose a new paradigm by forecasting prices by modeling \textit{investor behavior} in response to market information. \textsc{RoleCast} represents the market as investor archetypes, trains an LLM agent to simulate each archetype's portfolio actions from real-life 13F disclosures, and aggregates these actions into simulated market demand that can yield price forecasts through a closed-form equation. Imitating real investors poses two challenges: investor mandates are private, leaving no direct supervision for specifying agent roles, and the reasoning behind disclosed actions is unobservable, leaving only a sparse reward per decision. \textsc{RoleCast} tackles these in two stages: it searches generated candidate role personas for the one whose optimized agent best reproduces the archetype's disclosed decisions, then uses this agent as a teacher that grades every token of a student's own reasoning traces, providing dense feedback on its thinking process. Experiments on U.S. equities show that \textsc{RoleCast} outperforms state-of-the-art reasoning LLMs that forecast prices directly.
% , and we release our distilled investor agents to support research on financial market simulation. 
\vspace{-3px}
\end{abstract}



% \keywords{agent-based modeling, agent role learning, on-policy distillation}

%%
%% This command processes the author and affiliation and title
%% information and builds the first part of the formatted document.
\maketitle

% \begingroup
% \renewcommand{\thefootnote}{}%
% \footnotetext{\textit{Disclaimer.} This article is presented for research and
% informational purposes only and does not constitute investment advice or a
% recommendation to buy or sell any security or financial instrument. The results
% presented are based on simulated performance; simulated results have inherent
% limitations, and neither simulated nor past performance is indicative of future
% results.}%
% \endgroup

\begin{figure}[h]
\centering
\includegraphics[width=0.96\columnwidth]{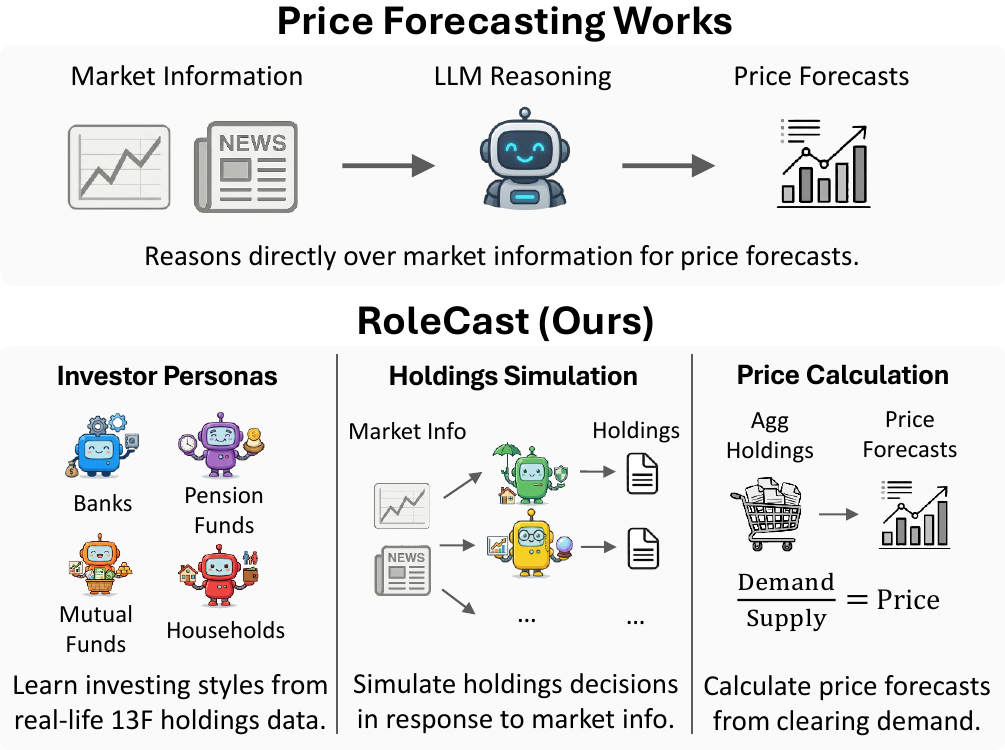}
\vspace{-6px}
\caption{Overview of the \textsc{RoleCast} framework. \textsc{RoleCast} models investor behaviors to simulate demand, then computes the clearing price forecasts via a closed-form equation.}
\label{fig:rolecast-overview}
\vspace{-8px}
\end{figure}

\section{Introduction} 
% With the advent of reasoning capabilities in Large Language Models (LLMs), many works have explored their use in financial forecasting \cite{lopez2023can, yang2023investlm, wang2023fingpt, xie2023pixiu}. The motivation of such works was to produce reasoning traces for stock forecasts, enabling investors to assess the underlying decision logic \cite{koa2024learning}. However, there is limited theoretical evidence to show that there exists a reliable way to ``reason'' over price movements. Financial literature has shown that even professional investors cannot forecast the market with certainty \cite{malkiel2019random}, and any exploitable advantage would be rapidly arbitraged away as argued by the Efficient Market Hypothesis \cite{fama1970efficient}.

Recent advances in the reasoning capabilities of Large Language Models (LLMs) have motivated their application to asset price forecasting \cite{liang2024fingpt, liu2025fin, li2026expred}. Existing methods typically prompt LLMs to infer future price movements directly from market information, often accompanied by reasoning traces for interpretability \cite{koa2024learning}. However, there is limited theoretical justification for treating price forecasting as a free-form reasoning problem \cite{xie2024finben, glasserman2024assessing}. Asset prices are equilibrium outcomes of heterogeneous investor decisions, while persistent predictive signals are likely to be rapidly incorporated into prices \cite{malkiel2019random, fama1970efficient}. This motivates a new question: \textit{rather than directly reasoning about prices, can LLMs model how investors respond to information and derive prices from their aggregated behavior?}

% At its core, a financial market aggregates the actions of participants with different beliefs, objectives, and institutional constraints \cite{mankiw1998principles}. We therefore propose \textsc{RoleCast}, a behavior-based forecasting framework that models the intermediate mechanism from market information to investor demand and, ultimately, to prices. As illustrated in Figure~\ref{fig}, \textsc{RoleCast} first partitions market participants into economically motivated investor archetypes \cite{chopra2024limits, koijen2019demand}. For each archetype, it learns an LLM agent that predicts portfolio decisions from market information using institutional holdings disclosed in SEC 13F filings \cite{sec13f}. The simulated decisions are then aggregated into market demand, from which price forecasts are derived through a closed-form, economically grounded market-clearing equation \cite{koijen2019demand}.

At its core, financial markets are driven by \textbf{the actions of market participants}, each making decisions on what price to buy an asset at \cite{mankiw1998principles}. 
% , based on available information and investment goals \cite{mankiw1998principles}. 
The same market information can produce different behavioral responses \cite{miller1977risk} (\eg a market downturn that causes a risk-averse bank to cut exposure may lead an opportunistic advisor to accumulate), and a single reasoning chain \cite{wang2023self} on price might not capture conflicting perspectives.  
Instead of reasoning about the impact of market information on prices, our work proposes a new paradigm by modeling the \textit{investor behavior} in response to market information. 
Specifically, we propose {\textsc{RoleCast}}, a framework that utilizes LLM agent \textbf{role} learning for price fore\textbf{cast}ing (see Figure \ref{fig:rolecast-overview}), which performs the following steps:

\vspace{2px} 
\noindent
\textbf{(1)} We split the universe of market participants into archetypes \cite{chopra2024limits}, where the type splits are informed by economic literature \cite{koijen2019demand}.

\vspace{2px} 
\noindent
\textbf{(2)} For each archetype, we teach an LLM agent to produce portfolio decisions in response to given information in the same way the investor type would invest, which is trained using real-world disclosure data from SEC 13F filings \cite{sec13f}. This can be seen as distilling \cite{li2025naturalthoughts} the behavior of real-world investors onto LLM agents. 

\vspace{2px} 
\noindent
\textbf{(3)} The simulated behaviors are mapped to price forecasts through an economic theory-grounded mechanism. Here, the portfolio actions for all investor types are aggregated to form the market demand \cite{koijen2019demand}, yielding a closed-form equation to obtain the forecasts.
% \vspace{2px}

% Learning investor behavior from portfolio disclosures can be formulated as an imitation-learning problem \cite{ho2016generative, ciosek2021imitation}, but presents two challenges. First, at the \textbf{role level}, different investor types should respond differently to the same information, yet their investment objectives and institutional constraints are generally unobserved. Second, at the \textbf{reasoning level}, disclosure data reveal portfolio outcomes but not the decision processes that produce them. Existing approaches often bootstrap reasoning with generated trajectories and outcome-level rewards \cite{guo2025deepseek, zelikman2024quiet}. Such scalar supervision provides limited credit assignment over long reasoning trajectories and does not indicate where an incorrect decision process diverges.

\begin{figure}[h]
\centering
\hspace{1px}
\includegraphics[width=0.9\columnwidth]{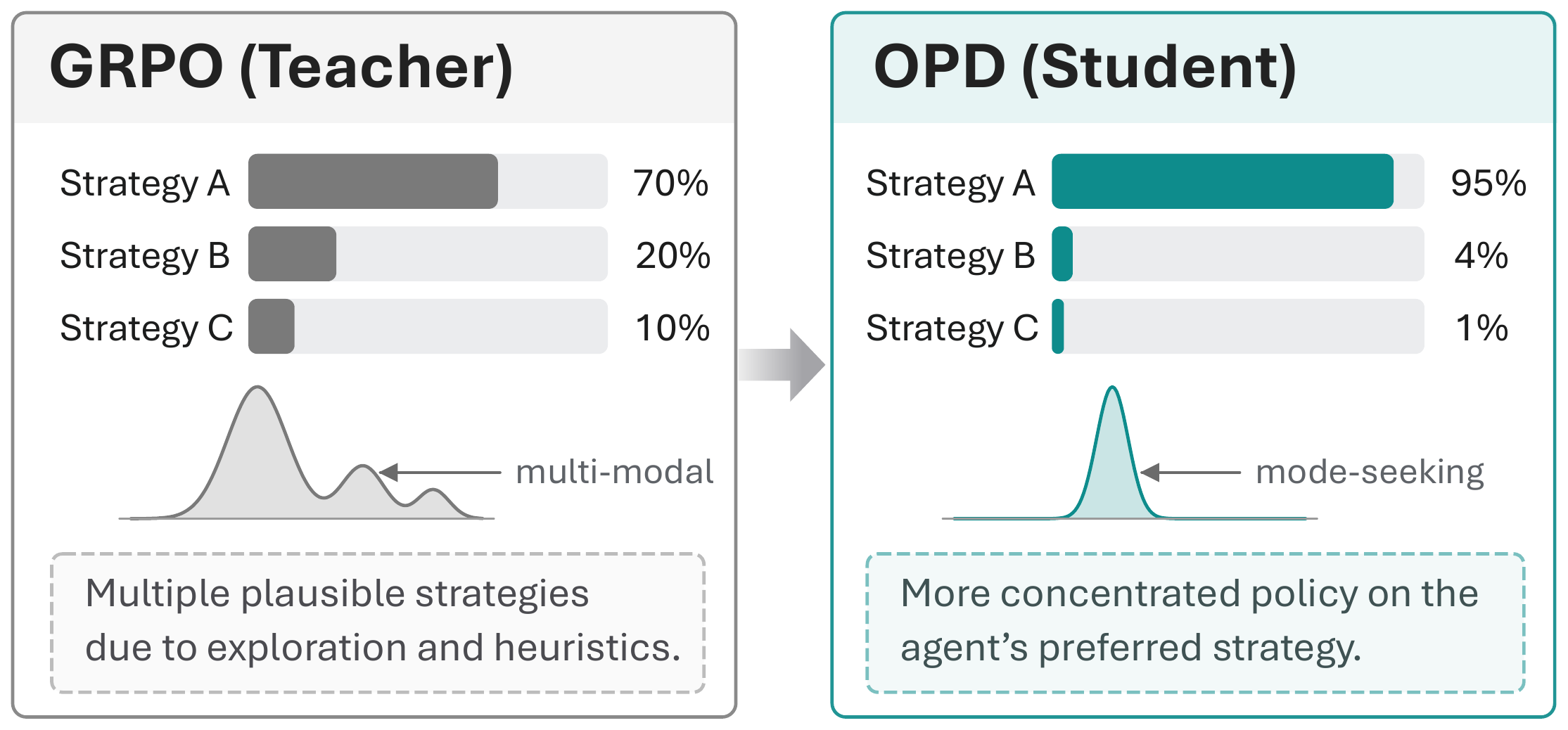}
\vspace{-6px}
\caption{Through on-policy distillation (OPD), the student model concentrates on
the teacher model's preferred strategy, allowing the distilled investor agents to imitate this behavior.}
\label{fig:opd-sharpening}
\end{figure}

The emulation of investment behavior is akin to an imitation learning problem \cite{ho2016generative, ciosek2021imitation}. 
For our task, we tackle two \textit{technical challenges}. 
At the \textbf{role-level}, agents representing different investor types should respond differently to the same market information. Such heterogeneity arises from differences in investment styles and institutional mandates, which are generally private and unavailable as direct training supervision. 
At the \textbf{thought-level}, while the model can learn investors' portfolio actions from SEC disclosure data, it does not see the reasoning processes behind these actions. 
Previous works have bootstrapped LLM training using generated reasoning traces from outcome-reward models \cite{guo2025deepseek, zelikman2024quiet}. 
% However, a sparse outcome reward compresses an entire reasoning trajectory into a single score, and a trajectory containing one mistaken step would receive the same feedback as a trajectory that was incorrect from the start. Thus, the agent might learn that its reasoning is erroneous, but is not able to learn which part of its thought process produced the mistake.
However, a sparse outcome reward assigns a single score to an entire reasoning trajectory, providing limited supervision over its intermediate steps. Consequently, the policy simply learns a distribution of plausible reasoning steps (see Figure \ref{fig:opd-sharpening}) that have worked well heuristically, but might not consistently reproduce the investors' strategy. 

% Figure 2 sits right after the paragraph that first cites it.

To address these challenges, we propose a two-stage imitation strategy, which combines agent role-learning \cite{shao2023character, salewski2023context, wang2026metagen} and on-policy distillation \cite{agarwal2024policy, ko2024distillm, gu2026minillm} techniques. 
To imitate \textbf{investor roles}, we first generate a library of candidate role personas from historical market information and investor-action pairs. 
Each persona instantiates an investor agent that reasons over the available market information and produces a portfolio action. 
The candidate agents are then optimized according to how closely their actions reproduce the disclosed decisions of the corresponding investor type \cite{shao2024deepseekmath}. 
We evaluate them on a held-out validation set, and retain the role persona that best reproduces the investor type's behavior. 
To imitate \textbf{investor thinking}, the behaviorally aligned agent is used as a teacher model, and a student model generates reasoning trajectories on-policy. 
Along each student-generated trajectory, the teacher then provides token-level distributional supervision, and the student is trained using a per-token reverse KL objective \cite{agarwal2024policy}. 
% Unlike a single outcome reward, this objective provides dense feedback throughout the reasoning trajectory, allowing the student to identify and correct errors within its reasoning process. 
The mode-seeking behavior \cite{gu2026minillm} of the objective \textit{sharpens} \cite{huang2025self} the learnt policy by concentrating the student on the teacher's preferred reasoning steps, which emphasizes their investing strategies.
The resulting student model serves as the distilled investor agent for the corresponding investor type in the \textsc{RoleCast} framework.

% To demonstrate the effectiveness of \textsc{RoleCast}, we perform extensive experiments and show that our framework is able to outperform state-of-the-art LLMs that perform direct price forecasting. 
We empirically verify the effectiveness of the \textsc{RoleCast} framework on stock prices, shares outstanding data from CRSP, and 13F holdings disclosure data from SEC, and show that \textsc{RoleCast} is able to outperform state-of-the-art LLMs that perform direct price forecasting by $+5.1\%$ on MAPE.
Through ablation studies, we also verify the usefulness of our agent role-learning and on-policy distillation components. 
% We release the fine-tuned investor agents to the public, which could be useful to research on stock market simulations. 
% \cite{yang2026twinmarket, li2025mars}. 
The main contributions of this paper are:
\begin{itemize}
[leftmargin=*]
% \vspace{-3px}
\item We propose \textsc{RoleCast}, a role-based agent simulation framework for behavior-driven stock price forecasting in response to market information.
% via an agent simulation framework utilizing role learning for forecasts.
\vspace{2px}
\item We devise a two-stage imitation strategy that learns archetype-specific investor roles from portfolio disclosures and distills their reasoning policies through dense on-policy supervision.
% To learn investor behavior, we tackle two technical challenges. A role-learning stage generates and selects role personas whose optimized agent best reproduces each investor type's actions, while an on-policy distillation process transfers the selected reasoning policy onto LLM investor agents through dense token-level supervision.
\vspace{2px}
\item  We empirically validate that our framework outperforms direct price forecasting methods, and show that the learnt personas and policies produce distinct, accurate investor holdings behaviors.
% We also release the finetuned agents to support research on market simulations.
\end{itemize}
\section{Related Work} 
\subsection{LLMs in Finance Forecasting}
% The advent of Large Language Models (LLMs) introduced many AI in finance works seeking to exploit their thinking capabilities on financial forecasting. The earliest works explored directly prompting LLMs for stock prediction using news headlines \cite{lopez2023can}; later works adapted general LLMs to the financial domain, either by pre-training on financial corpora \cite{wu2023bloomberggpt} or instruction-tuning open-source models \cite{yang2023investlm, wang2023fingpt, xie2023pixiu}. Subsequent works further produced reasoning traces over stock tweets \cite{koa2024learning, li2026expred} and financial time-series \cite{yu2023harnessing, koa2025reasoning}. Across these works, the LLM typically reasons directly on market information to forecast future prices. In contrast, our work seeks to reproduce investors' reasoning on market information, then forecasts the market price from their aggregated demand.
The advent of Large Language Models (LLMs) introduced many AI in finance works seeking to exploit their thinking capabilities on asset price forecasting.
Early works explored directly prompting LLMs for price predictions using news headlines \cite{lopez2023can}; later works adapted general LLMs to the financial domain, by pre-training on financial corpora \cite{wu2023bloomberggpt} or instruction-tuning open-source models \cite{yang2023investlm, wang2023fingpt, xie2023pixiu}. Subsequent works further produced reasoning traces over stock tweets \cite{koa2024learning, li2026expred} and financial time-series \cite{yu2023harnessing, koa2025reasoning}. These works typically reason directly over market information to forecast prices. In contrast, our work seeks to reproduce investors' reasoning on information, then forecast the price from their aggregated demand.

\subsection{Agent Role-Learning}
% Prior work has found that LLMs can adopt roles when given a persona prompt, which measurably alters their behavior \cite{salewski2023context}. On the character simulacra task \cite{park2023generative, xie2025human}, works have explored reconstructing role profiles from existing sources, such as biographical experiences \cite{shao2023character}, film scripts \cite{wang2024rolellm} and encyclopedic character knowledge \cite{lu2024large}. Unlike these use cases, it is difficult to obtain behavioral source texts that inform investors' portfolio strategies.
% for our task
% In parallel, multi-agent frameworks also assign roles to LLM agents for complex task completion (\eg software development \cite{hong2024metagpt}), either through handcrafted prompts \cite{wu2023autogen, qian2024chatdev} or role self-evolution \cite{wang2026metagen}. In contrast, the LLM agents in our work are \textit{not} optimized for how well they cooperate to complete a task, but rather for how well they can simulate role behavior.  
Prior work has found that LLMs can adopt roles when given a persona prompt, which measurably alters their behavior \cite{salewski2023context}. Character simulacra works \cite{park2023generative, xie2025human} reconstruct role profiles from existing sources such as biographies \cite{shao2023character}, film scripts \cite{wang2024rolellm} and encyclopedic character knowledge \cite{lu2024large}. For our use case, it is difficult to obtain behavioral source texts for investors' strategies. Multi-agent frameworks \cite{hong2024metagpt} also assign roles to LLM agents through handcrafted prompts \cite{wu2023autogen, qian2024chatdev} or role self-evolution \cite{wang2026metagen}. In contrast, our agents are optimized not for how well they cooperate on a task, but how faithfully they can simulate actual role behavior.

\begin{figure*}[ht]
\centering
\includegraphics[width=0.93\textwidth]{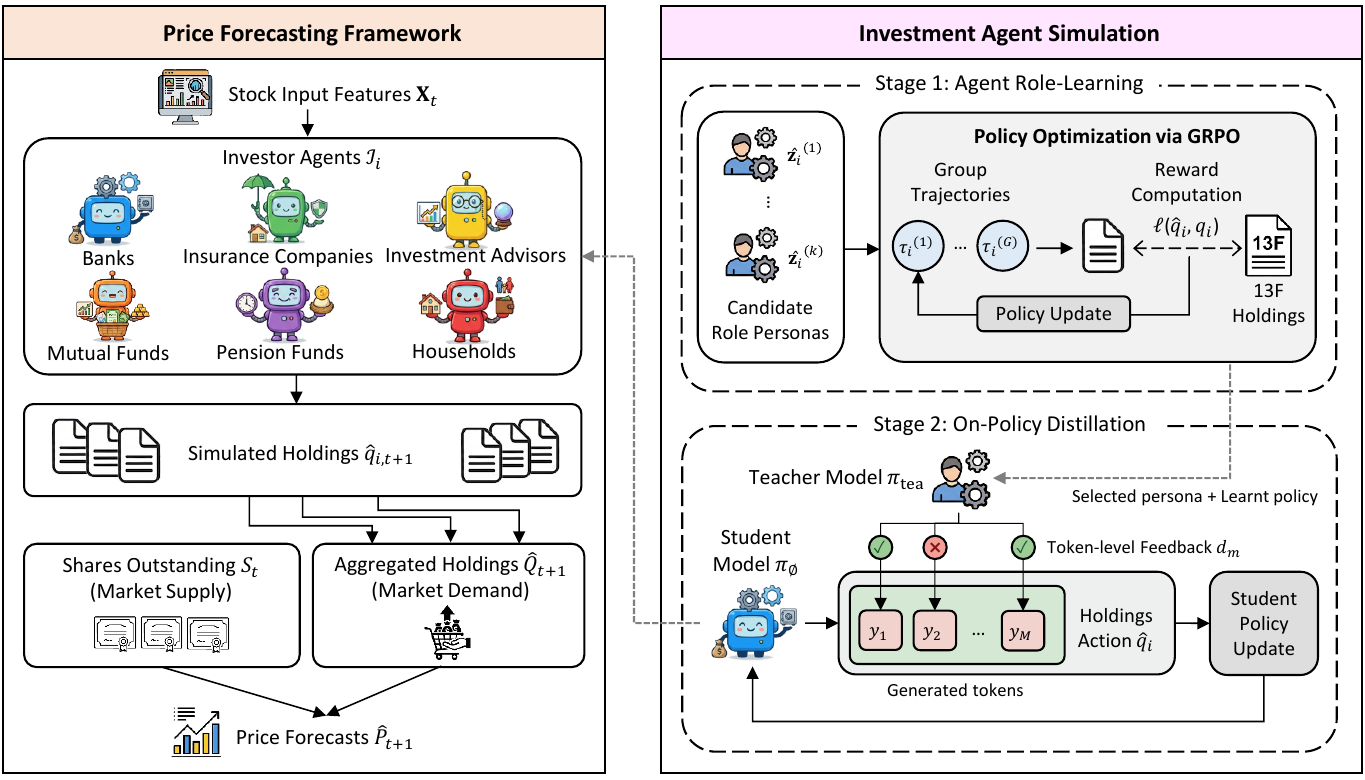}
\vspace{-5px}
\caption{Visualization of the \textsc{RoleCast} framework, which simulates actual market dynamics based on holdings actions from investor agents. We distill actual investor behavior into these agents through role-learning and on-policy distillation.}
\vspace{-6px}
\label{fig:rolecast-framework}
\end{figure*}

\subsection{On-Policy Distillation}
% Knowledge distillation for LLMs was generally performed off-policy, by fine-tuning the student on teacher-generated outputs \cite{hinton2015distilling, li2025naturalthoughts}. This trains the student on states visited by the teacher rather than its own, causing errors to compound at inference \cite{agarwal2024policy}. Conversely, on-policy reinforcement learning trains on the student's own samples \cite{guo2025deepseek, shao2024deepseekmath}, but returns a single reward for the entire trajectory, offering no signal on which part of the reasoning was wrong. On-policy distillation (OPD) combines the strengths of both techniques by having a teacher grade every token of the student's own samples \cite{gu2026minillm} through a reverse KL-divergence objective \cite{ko2024distillm, ko2025distillm}. These works typically assume an existing competent teacher model; in our work, we first search for the most representative teacher models by utilizing candidate persona rollouts. 
Knowledge distillation for LLMs was generally performed off-policy, by fine-tuning the student on teacher-generated outputs \cite{hinton2015distilling, li2025naturalthoughts}. This trains the student on states visited by the teacher rather than its own, causing errors to compound at inference \cite{agarwal2024policy}. Conversely, on-policy reinforcement learning trains on the student's own samples \cite{guo2025deepseek, shao2024deepseekmath}, but its single trajectory-level reward gives no signal on which reasoning step was wrong. On-policy distillation (OPD) combines both: a teacher grades every token of the student's outputs \cite{gu2026minillm} through a reverse KL-divergence objective \cite{ko2024distillm, ko2025distillm}. These works assume a competent teacher exists; we instead search for the most representative teacher via candidate persona rollouts.
\section{The \textsc{RoleCast} Framework} 
In this section, we discuss the overall \textsc{RoleCast} framework (see Figure \ref{fig:rolecast-framework}), and the proposed agent role-learning and on-policy distillation components for learning investor actions.

\subsection{Price Forecasting Framework} 
For each stock at each quarter $t$, we observe input $\mathbf{X}_t = \left(\mathbf{p}_t; \mathbf{x}_t^{\mathrm{text}}\right)$, where $\mathbf{p}_t$ denotes its historical price series and $\mathbf{x}_t^{\mathrm{text}}$ denotes textual information from quarterly reports. Our objective is to forecast its price $\hat{P}_{t+1}$ in the subsequent quarter $t+1$. While existing approaches typically learn a direct forecasting function, $\hat{P}_{t+1} = \mathcal{F}(\mathbf{X}_t)$, we instead design an agent simulation framework to model the market dynamics.

To realistically simulate a universe of all real-life investors, we first split them into $N$ investor archetypes \cite{chopra2024limits}. Each type represents a group of investors with similar investing behavior (\eg mutual funds might be driven by profitability while banks might be driven by capital preservation) and the type splits are informed by economic literature \cite{koijen2019demand}. Each investor type is modeled by an LLM agent, denoted by $\{\mathcal{I}_i\}_{i=1}^N$. 

At each time $t$, the stock holdings $q$ of these investors can be observed using their 13F filings, which are mandatory disclosures for investors with substantial stock holdings and are publicly available data \cite{sec13f}. These filings allow us to observe the investment behavior of each investor type.

For each investor-type LLM agent $\mathcal{I}_i$, the aim is to produce simulated holdings that reflect the realistic behavior of its corresponding investor group. Specifically, the agent simulates the stock holdings of investor type $i$ for the next time step,
\begin{equation}
\hat{q}_{i,t+1} = \mathcal{I}_i(\mathbf{X}_t).
\end{equation}
The agents would be rewarded based on how well their simulated holdings $\hat{q}_{i,t+1}$ align with actual observed holdings.

Finally, to get price forecasts, the simulated holdings are aggregated to get the overall market demand for each stock, 
\begin{equation}
\hat{Q}_{t+1} = \sum_{i=1}^N \hat{q}_{i,t+1}.
\end{equation}
The price forecast is then obtained by scaling the simulated \textit{demand} by the  available \textit{supply} of shares on the market, which is provided by the number of shares outstanding $S_t$:
\begin{equation}
\hat{P}_{t+1} = \frac{\hat{Q}_{t+1}}{S_t}.
\label{price_mechanism}
\end{equation}
Intuitively, the price of a stock would increase when the forecasted demand is high relative to the available supply. 
% A formalized proof of this relation is provided in Appendix \ref{price_formation}.

\subsection{Stage 1: Agent Role-Learning}
The core of the  \textsc{RoleCast} framework relies on simulating realistic investing behavior in its agents. This is done by searching for the optimal role personas, then distilling policies onto agents to simulate investor behaviors at inference.

Formally, each investor type can be represented as an agent $\mathcal{I}_i$ with a unique role persona $\mathbf{z}_i$. At each time $t$, the agent would observe the input features $\mathbf{X}_t$ of a stock and reason about how much of the stock to hold. This is represented by a reasoning trace $\mathbf{r}_{i, t+1}$, and the holdings decision $q_{i, t+1}$. Thus, the decision generation process can be visualized as:
\begin{equation}
\mathbf{z}_i \rightarrow \mathbf{r}_{i,t} \rightarrow q_{i,t}.
\end{equation}

However, while $q_{i,t+1}$ can be observed during training, there are no gold-standard labels for the investors' investment styles or their actual reasoning process for each decision.

At the role-level, each persona $\mathbf{z}_i$ can be defined as a tuple: 
\begin{equation}
\mathbf{z}_i \;=\; \left(V_i,\; D_i,\; R_i,\; U_i\right),
\label{eq:role}
\end{equation}
where $V_i$ is the investor-type label (\eg \textit{mutual fund} or \textit{bank}), $D_i$ is the semantic description of the persona, $R_i$ is the system-level prompt template instantiated with $V_i$, and $U_i$ is the user-level template describing the required task. 

To learn the role personas, we first generate a set of candidate role personas $\{\hat{\mathbf{z}_i}^{(k)}\}_{k=1}^{K}$. This is done by letting a pretrained LLM reason over $K$ demonstration sets of input-output pairs $\left(\mathbf{X}_t, q_{i,t+1}\right)$ drawn from the train split, producing a brief description that characterizes the investment style.

Each candidate instantiates a role-conditioned agent which learns a policy $\pi_\theta(\hat{q}_{i,t+1} \mid \hat{\mathbf{z}_i}^{(k)}, \mathbf{X}_t)$. Here, we utilize the reward function in GRPO fine-tuning \cite{shao2024deepseekmath} to enforce that the simulated holdings would match the investors' behavior. We let $\boldsymbol{\tau}_{i,t}=\{\hat{\mathbf{r}}_{i,t}, \hat{q}_{i,t}\}$ be the generated trajectory from the agent, consisting of the simulated reasoning trace $\hat{\mathbf{r}}_{i,t}$ and holdings $\hat{q}_{i,t}$. We minimize the following objective:
% \vspace{-5px}
\begin{multline}
\mathcal{L}_{\text{GRPO}}(\theta_i)
=
\\
-
\mathbb{E}
\Bigg[
\frac{1}{G}
\sum_{g=1}^G
\Bigg(
\min \Bigg(
\frac{
\pi_{\theta_i}(\boldsymbol{\tau}_{i,t+1}^{(g)} \mid \mathbf{x}_t, \mathcal{Z}_i)
}{
\pi_{\theta_{\text{old}}}(\boldsymbol{\tau}_{i,t+1}^{(g)} \mid \mathbf{x}_t, \mathcal{Z}_i)
}
A_i^{(g)},
\Bigg)
\Bigg].
\vspace{-5px}
\end{multline}
$A_i^{(g)}$ denotes the advantage learnt by the LLM policy for each group $G$, which is derived from a set of rewards $\mathcal{R}$:
\begin{equation}
A_i^{(g)}
=
\frac{
\mathcal{R}_i^{(g)}
-
\text{mean}\!\left(\left\{ \mathcal{R}_i^{(1)}, \dots, \mathcal{R}_i^{(G)} \right\}\right)
}{
\text{std}\!\left(\left\{ \mathcal{R}_i^{(1)}, \dots, \mathcal{R}_i^{(G)} \right\}\right)
}.
\end{equation}
The reward function is defined as:
\begin{equation}
\mathcal{R}_i
=
\lambda_r \,
\mathbb{I}\!\left[
\{\hat{\mathbf{r}}_{i,t} \subset \boldsymbol{\tau}_{i,t}
\right]
-
\lambda_q \,
\ell\!\left(
\hat{q}_{i,t}, q_{i,t}
\right).
% -
% \lambda_z \,
% \ell\!\left(
% \hat{\mathbf{z}}_{i}, \mathbf{z}_{i}
% \right).
\end{equation}
where $\mathbb{I}(\cdot)$ enforces reasoning format \cite{guo2025deepseek}, and distance metric $\ell(\cdot)$ penalizes deviations between the predicted holdings $\hat{q}_{i,t}$ and the ground-truth value.

% the model to always yield a reasoning process $\hat{\mathbf{r}}_{i,t}$ before answering
% Using , t
% The reward function , and that 

\paragraph{Role selection.}
Because all candidates of a type share $R_i$ and task prompt $U_i$, they differ only in the persona, so downstream performance differences are attributable to the role. Here, each optimized candidate $\theta_i^{(k)}$ is evaluated on the
held-out validation split of its investor type, and the role is selected by the distance between its simulated and disclosed
holdings:
\begin{equation}
k_i^{\ast} \;=\; \operatorname*{arg\,min}_{k \in \{1,\ldots,K\}} \;
\mathbb{E}
\left[ \ell\left(\hat{q}_{i,t}^{(k)},\, q_{i,t}\right) \right],
\label{eq:role-select}
\end{equation}
% where $\hat{q}_{i,t}^{(k)}$ denotes the holdings simulated by candidate $\theta_i^{(k)}$. 
We retain the best role persona $\hat{\mathbf{z}_i}^{(k_i^{\ast})}$ and its optimized policy as the behaviorally-aligned agent for investor type $i$, which then serves as the teacher model in the distillation stage.

\subsection{Stage 2: On-Policy Distillation}

Although Stage 1 learns behaviorally aligned policies through GRPO, its supervision remains trajectory-level: each generated reasoning trajectory and investment decision receives only a single scalar reward. 
Such outcome-level supervision provides no direct indication of which intermediate reasoning steps are responsible for an incorrect decision. To provide finer-grained supervision, we further distill the role-conditioned policy learned in Stage 1 into a student policy using on-policy distillation \cite{li2026videoopd}.
Specifically, the optimized policy obtained after role learning is frozen as the teacher policy $\pi_{\mathrm{tea}}$, while $\pi_{\phi}$ denotes the student policy to be optimized.
% Role learning produces a behaviorally aligned agent for each investor type, but its training signal remains sparse: the GRPO reward grades the entire trajectory with a single scalar score, which cannot indicate which part of the reasoning produced an error. To supervise the thinking process itself, we distill the selected agent into a student agent through on-policy distillation \cite{li2026videoopd}, which converts this outcome-level feedback into a dense, token-level learning signal. The optimized policy from role selection serves as the teacher $\pi_{\mathrm{tea}}$, to be distilled into a student policy $\pi_{\phi}$.
Given input features $\mathbf{X}_t$ and role persona $\mathbf{z}_i$, the student model first samples a single trajectory from its current policy,
\begin{equation}
\boldsymbol{\tau}_{i,t} = \left(y_1, \ldots, y_M\right) \sim \pi_{\phi}\left(\cdot \mid \mathbf{X}_t, \mathbf{z}_i\right),
\label{eq:opd-sample}
\end{equation}
where $\boldsymbol{\tau}_{i,t}$ is the token-level sequence realizing the reasoning trace and holdings decision $\{\hat{\mathbf{r}}_{i,t}, \hat{q}_{i,t}\}$, and $M$ is the token length. 
Because trajectories are drawn from the student itself, the training states match the states the student induces at inference, avoiding the compounding errors of off-policy distillation on teacher-generated text \cite{agarwal2024policy}.
For each student-generated token, the teacher computes the conditional log-probability $\log \pi_{\mathrm{tea}}(y_m \mid \mathcal{S}_m)$, where $\mathcal{S}_m = (\mathbf{X}_t, \mathbf{z}_i, y_{<m})$ denotes the generation state. 
Here, the teacher is used only to evaluate the student's samples, ensuring strict on-policy optimization.  

Together with the student's log-probabilities over the same tokens, we define a dense per-token reward that is induced by the reverse Kullback–Leibler (KL) divergence:
\begin{equation}
d_m = -\big( \log \pi_{\phi}\left(y_m \mid \mathcal{S}_m\right) - \log \pi_{\mathrm{tea}}\left(y_m \mid \mathcal{S}_m\right) \big).
\label{eq:opd-reward}
\end{equation}
The tokens which the teacher assigns low probability incur large penalties, so $d_m$ pinpoints the exact positions where the student's reasoning deviates from the teacher's, rather than the trajectory as a whole. 
In expectation, the negated reward at each state equals the reverse KL between the two policies,
\begin{equation}
\mathbb{E}\big[ -d_m \big]
= D_{\mathrm{KL}}\big( \pi_{\phi}(\cdot \mid \mathcal{S}_m) \,\big\|\, \pi_{\mathrm{tea}}(\cdot \mid \mathcal{S}_m) \big),
\label{eq:opd-kl}
\end{equation}
so maximizing the rewards along the student's own trajectories minimizes the reverse KL at every state the student visits. 
This objective is mode-seeking: the student concentrates on the teacher's preferred decision paths instead of covering all of its low-probability variants \cite{gu2026minillm}.
 
The student model is then updated with a standard policy gradient, treating $d_m$ as a token-level feedback reward:
\begin{equation}
\mathcal{L}_{\mathrm{OPD}}(\phi)
= - \mathbb{E} \Bigg[ \sum_{m=1}^{M} d_m \, \nabla_{\phi} \log \pi_{\phi}(y_m \mid \mathcal{S}_m) \Bigg],
\label{eq:opd-update}
\end{equation}
where $d_m$ is treated as a constant with respect to $\phi$. 

% \paragraph{Role selection.}
The finetuned student model for each investor type $i$ serves as the distilled investor agent $\mathcal{I}_i$, and the simulated holdings $\hat{q}_{i,t+1}$ across the $N$ distilled agents are then aggregated into price forecasts $\hat{P}_{t+1}$ through the \textsc{RoleCast} framework.

% does not generate its own tokens: it 
\section{Experiments}
 
\subsubsection{Dataset.}
We evaluate \textsc{RoleCast} on the universe of the S\&P 500 constituent stocks. The duration of the data ranges across 10 years, from 2014Q2 to 2024Q3. The data on stock prices and shares outstanding are collected from the Center for Research in Security Prices (CRSP) Database. The institutional stock holdings are taken from the Thomson Reuters Institutional Holdings Database, which are compiled from publicly-available SEC 13F disclosure filings \cite{sec13f}.  

At each quarter $t$, the input $\mathbf{X}_t$ for each stock consists of: (1) the textual information $\mathbf{x}_t^{\mathrm{text}}$, given as the Management's Discussion and Analysis (MD\&A) section of the latest 10-K and 10-Q filings from SEC EDGAR; and (2) the historical price series $\mathbf{p}_t$, given as the daily close prices, trading volumes and log returns in the last quarter from Yahoo Finance. 

To prevent look-ahead bias, all features are also subject to an information cut-off of 45 days after each quarter-end (\ie the 13F disclosure deadline), such that the agents would only observe information that was already released to public when the corresponding holdings decisions were disclosed. 

% More details on look-ahead bias control are in Appendix \ref{app:leakage}.

% Layout: -3px (was -2px) before this and the next two run-in headings, so the
% left column of page 5 ends exactly with the first `Parameter Settings' paragraph.
\vspace{-3px}
\subsubsection{Baselines.}
We compare against five categories of works.

\vspace{2px} 
\noindent
(1) \textbf{Na\"ive}: Random Walk \cite{malkiel2019random}, which our work simplifies to carrying the last observed price forward.

\vspace{2px} 
\noindent
(2) \textbf{Zero-shot LLMs (General)}: We compare with Mistral-7B-Instruct-v0.3 \cite{jiang2023mistral}, Llama-3.1-8B-Instruct \cite{grattafiori2024llama} and Qwen2.5-7B-Instruct \cite{yang2024qwen25}, which is also our base model and serves as direct comparison between forecasting price and holdings (\textsc{RoleCast}).

\vspace{2px} 
\noindent
(3) \textbf{Zero-shot LLMs (Finance)}: Fino1 \cite{qian2025fino1}, Fin-R1 \cite{liu2025fin}, which are trained for financial reasoning; and Llama-Fin \cite{ke2025demystifying}, Qwen-Open-Finance \cite{caillaut2025llm}, which are used in financial forecasting.

\vspace{2px} 
\noindent
(4) \textbf{Fine-tuned LLMs}: We fine-tuned selected LLMs from the above which are open-sourced on our training data, to directly learn the forecasting function $\hat{P}_{t+1} = \mathcal{F}(\mathbf{X}_t)$.

\vspace{2px} 
\noindent
(5) \textbf{Time-series LLMs}: ChatTS \cite{xie2024chatts}, ChatTime \cite{wang2025chattime}, TimeCAP \cite{lee2025timecap} and TimeOmni \cite{guan2026timeomni}, which adapt LLMs to analyze numerical time-series inputs. Among these models, ChatTS and TimeCAP are finetuned for the forecasting task. 

\vspace{-3px}
\subsubsection{Evaluation Metrics.}
We evaluate across four metrics, all reported in percentages (lower is better). MAPE averages the per-sample relative errors, weighting every stock equally. WAPE normalizes the total absolute error by the total true price, so high-priced stocks carry more weight and errors on low-priced stocks are not inflated. sMAPE normalizes each error by the average magnitude of the true and predicted values, making it symmetric between over- and under-prediction. MAE reports the average absolute error in dollars.

\vspace{-3px}
\subsubsection{Parameter Settings.}
All investor agents in \textsc{RoleCast} are trained using Low-Rank Adaptation (LoRA) \cite{hu2022lora}, with Qwen-2.5-7B Instruct as the base model. For agent role learning, we generate candidate personas by prompting GPT-5.4-mini  with 10 sampled demonstration pairs from the train split. Each candidate persona is then optimized with GRPO for 500 steps, using $G=4$ rollouts per prompt at temperature 0.9, a learning rate of $1\mathrm{e}{-6}$, and a KL coefficient of 0.02. The distance function $l$ used is the Huber loss \cite{huber1992robust}.

For on-policy distillation, the student is trained using LoRA on the same base model, which samples one trajectory per prompt at temperature 0.9 and top-$p = 0.95$, while the frozen teacher scores its tokens. The student is trained with the reverse KL policy gradient for 300 steps, using AdamW with a learning rate of $1\mathrm{e}{-5}$ and 8 prompts per train step.

We define $N=7$ investor agent types: \textit{Banks}, \textit{Insurance Companies}, \textit{Investment Advisors}, \textit{Mutual Funds}, \textit{Pension Funds}, \textit{Households} and \textit{Others}, which were used as investor types in economics literature \cite{koijen2019demand}. 

% An audit of the agent type split can be found in Appendix \ref{app:audit}.
\section{Results}

\begin{table}[t]
\small
\centering
\caption{Performance comparison. The best baselines are \underline{underlined}, and the best results are \textbf{bolded}.}
\vspace{-6px}
% \resizebox{1.0\columnwidth}{!}{
\begin{tabular}{lcccc}
\toprule
\textbf{Model} & \textbf{MAPE} & \textbf{WAPE} & \textbf{sMAPE} & \textbf{MAE} \\
\midrule
\rowcolor{gray!20}\multicolumn{5}{c}{\textbf{\textit{Na\"ive}}} \\
Random Walk                            & 11.02 & 11.43 & 11.29 & 23.50 \\
\rowcolor{gray!20}\multicolumn{5}{c}{\textbf{\textit{Zero-shot LLMs (General)}}} \\
Mistral             & 12.50 & 12.10 & 12.99 & 22.11 \\
Llama  & 11.59 & 12.01 & 11.71 & 24.09 \\
Qwen                                & 10.71 & 10.96 & 10.91 & 22.54 \\
\rowcolor{gray!20}\multicolumn{5}{c}{\textbf{\textit{Zero-shot LLMs (Finance)}}} \\
Fino1             & 13.84 & 12.71 & 16.69 & 25.88 \\
Fin-R1  & 13.30 & 12.56 & 15.21 & 21.97 \\
Llama-Fin    & 12.54 & 11.72 & 13.50 & 22.88 \\
Qwen-Open-Fin          & 13.08 & 12.20 & 13.32 & 21.82 \\
\rowcolor{gray!20}\multicolumn{5}{c}{\textbf{\textit{Fine-tuned LLMs}}} \\
Mistral                                & 10.65 & 11.23 & 10.80 & 22.34 \\
Llama                                  & 11.72 & 11.83 & 12.28 & 24.33 \\
Qwen                                & 10.78 & 10.98 & 10.91 & 22.58 \\
Fino1             & 11.01 & 11.17 & 11.39 & 22.75 \\
Fin-R1               & 11.02 & 11.36 & 11.09 & 23.39 \\
\rowcolor{gray!20}\multicolumn{5}{c}{\textbf{\textit{Time-series LLMs}}} \\
ChatTS            & 14.39 & 17.35 & 11.56 & 35.76 \\
ChatTime       & \underline{10.57} & \underline{10.31} & \underline{10.40} & \underline{21.23} \\
TimeCAP         & 12.41 & 12.00 & 12.44 & 24.68 \\
TimeOmni       & 16.91 & 17.25 & 16.66 & 35.50 \\
\midrule
\textsc{RoleCast} (Ours)                             & \textbf{10.03} & \textbf{10.24} & \textbf{9.91} & \textbf{21.05} \\
\midrule
\textbf{Improvement} & $+5.1\%$ & $+0.7\%$ & $+4.7\%$ & $+0.8\%$ \\
\bottomrule
\end{tabular}
% }
\label{tab:main_results}
\vspace{-11px}
\end{table}

\subsubsection{Overall Performance.}
Table \ref{tab:main_results} reports the price forecasting results across all baselines. We can observe the following:
\begin{itemize}
[leftmargin=*]
\item The na\"ive Random Walk remains a competitive baseline, outperforming most of the zero-shot LLMs and matching several of the fine-tuned ones. This is consistent with the Efficient Market Hypothesis \cite{fama1970efficient}: most exploitable information is already reflected in the current price, leaving little consistent signal for a model to ``reason'' over. This observation supports our original motivation of forecasting prices through investor simulation.

\item Among the general zero-shot LLMs, Qwen performs the best, which validates its choice as the backbone model.

\item Among the finance zero-shot LLMs, the better performing ones are Llama-Fin and Qwen-Open-Finance, which are originally finetuned for the financial forecasting task.

\item Fine-tuning the LLMs on the forecasting task improves most of them, but all fine-tuned variants converge into a narrow band around the Random Walk level. Notably, the finance models lose their disadvantage and gain no edge over the general LLMs after fine-tuning. This suggests that learning a direct forecasting function $\mathcal{F}(\mathbf{X}_t)$ hits a ceiling once the persistence component of prices is captured, with limited further signals to be extracted.

\item Among the time-series LLMs, ChatTime is the strongest baseline overall, likely due to forecasting being its native task, unlike other models that only perform reasoning.

\item Finally, \textsc{RoleCast} outperforms the strongest baseline (ChatTime) across all four metrics. Since \textsc{RoleCast} is built on the same backbone as the fine-tuned Qwen baseline and performed better, the gains could be attributable to the framework itself rather than to model capacity. The improvements are also largest on the equally-weighted relative metrics (+5.1\% MAPE, +4.7\% sMAPE), indicating that the gains are broad-based across the stock universe rather than driven by a few high-priced names.
\end{itemize}

% Statistical significance of the results are given in Appendix \ref{app:significance}.

\begin{table}[h]
\vspace{-3px}
\caption{Ablation study of agent roles for \textsc{RoleCast}.}
\vspace{-8px}
\centering
\small
\begin{tabular}{lcccc}
\toprule
\textbf{Variant} & \textbf{MAPE} & \textbf{WAPE} & \textbf{sMAPE} & \textbf{MAE} \\
\midrule
No persona        & \underline{10.95} & \underline{11.74} & \underline{11.02} & \underline{24.13} \\
Worst persona     & 14.57 & 13.35 & 11.26 & 27.45 \\
Random persona & 21.31 & 18.81 & 11.33 & 38.68 \\
\midrule
Learnt persona (Ours)           & \textbf{10.03} & \textbf{10.24} & \textbf{9.91} & \textbf{21.05} \\
\bottomrule
\end{tabular}
\label{tab:ablation_role}
\vspace{-7px}
\end{table}

\paragraph{Ablation on Agent Role Learning.}
Table \ref{tab:ablation_role} reports the performance across different role persona learning variants:
\begin{itemize}
[leftmargin=*]
\item \textit{No persona} removes the role description, such that each agent is optimized with the same GRPO procedure but conditioned only on the shared task prompts. Its performance falls to the level of the directly fine-tuned LLMs in Table \ref{tab:main_results}. Without explicit role personas, the agents drift towards similar average behaviors, and aggregating their holdings provides little benefit over a direct forecaster. 
% This shows that the role personas induce the heterogeneous responses needed for a meaningful demand simulation.

\item \textit{Worst persona} selects the candidate whose agent performed worst on the validation split (\ie taking $\arg\max$ instead of $\arg\min$ in the role selection step). Performance degrades despite the identical GRPO optimization, showing that fine-tuning cannot fully compensate for a misspecified role. As all candidates of a type share the same task prompts, this gap is likely attributable to the persona itself, which validates the need for the selection stage.

\item \textit{Random persona} assigns an arbitrary persona that is not generated from the investor type's demonstration pairs, and produces the poorest results. This shows that an ungrounded persona would actively misdirect the agent's reasoning. We also note that sMAPE stays flat across the degraded variants while MAPE and MAE inflate substantially; as sMAPE saturates under large over-predictions while MAPE is unbounded, this suggests that misspecified personas cause the agents to systematically over-estimate their holdings, inflating the simulated demand.

\item \textit{Learnt persona (Ours)} achieves the best results across all metrics, improving over the strongest variant \textit{No persona} by 8.4\% on MAPE. This validates both parts of the role-learning stage: grounding the candidate personas in real disclosure data, and selecting the persona whose optimized agent best reproduces the investor type's behavior.
\end{itemize}

\begin{table}[h]
\caption{Ablation study of OPD for \textsc{RoleCast}.}
\vspace{-8px}
\centering
\small
\begin{tabular}{lcccc}
\toprule
\textbf{Variant} & \textbf{MAPE} & \textbf{WAPE} & \textbf{sMAPE} & \textbf{MAE} \\
\midrule
Zero-shot             & 69.25 & 92.96 & 42.53 & 191.16 \\
% Zero-shot (with profile)           & 349.97 & 360.84 & 89.97 & 741.98 \\
OP-FKD & 15.84 & 14.70 & 10.59 & 30.23 \\
OP-RKD & 18.50 & 25.22 & 10.45 & 51.87 \\
GRPO                               & \underline{12.40} & \underline{12.17} & \underline{10.11} & \underline{25.02} \\
\midrule
OPD (Ours)                         & \textbf{10.03} & \textbf{10.24} & \textbf{9.91} & \textbf{21.05} \\
\bottomrule
\end{tabular}
\label{tab:ablation_opd}
\vspace{-8px}
\end{table}

\paragraph{Ablation on On-Policy Distillation.}
Table~\ref{tab:ablation_opd} compares OPD against alternative distillation methods. We observe:
\begin{itemize}
[leftmargin=*]
    \item \textit{Zero-shot} prompts the base model with the selected role persona, without fine-tuning. Its errors are far larger than all trained variants, showing that prompting personas alone cannot produce calibrated holdings values. 
    % the behavior of each agent must be learnt from the holdings disclosure data. 
    % This contrasts with the zero-shot LLMs in Table~\ref{tab:main_results}, which remain within a reasonable range when forecasting prices directly (likely anchored by the recent price levels in the input), highlighting that behavior simulation is the harder task.
    
    \item  In Off-Policy Forward KL Distillation (\textit{OP-FKD}), the student model samples its own trajectories but is trained to match the teacher's full token distribution. The forward KL is mass-covering: it pushes the student to spread probability over all of the teacher's plausible variants, including its low-probability reasoning paths, which dilutes the overall precision of the final holdings decisions.
    
    \item Off-Policy Reverse KL Distillation (\textit{OP-RKD}) instead minimizes the reverse KL score directly as a differentiable token-level loss, rather than using it as a reward. It underperforms OP-FKD on most metrics; a possible reason is that the mode-seeking objective is harder to optimize as a direct loss, where the student can collapse onto narrow reasoning patterns before its numeric outputs can be well-calibrated. This is also reflected in its large WAPE and MAE relative to its sMAPE, indicating that its errors are mostly concentrated in a subset of large mistakes.
    
    \item \textit{GRPO} uses the teacher model directly as the investor agent, without the distillation stage. It is the strongest variant, but its trajectory-level reward cannot indicate which part of a reasoning trace produced an error, which limits how well the thinking process itself can be optimized.
    
    \item \textit{OPD (Ours)} combines the strengths of the above: trajectories are sampled from the student itself, while the reverse-KL reward provides dense token-level feedback for each trajectory. It achieves the best results across all metrics. Notably, the distilled student agent outperforms its teacher model: It is possible that the mode-seeking objective concentrates the student on the teacher's preferred decision paths while discarding its noisier exploratory behaviors, effectively de-noising the final learnt policy.
    % \vspace{2px}
\end{itemize}

% \begin{figure}[]
% \centering
% \includegraphics[width=\columnwidth]{images/fig_holdings_price_scatter.pdf}
% \caption{Holdings accuracy vs.\ price accuracy per stock-quarter for \textsc{RoleCast}, under (a) equal-weighted (MAPE) and (b) dollar-weighted (WAPE) holdings error.}
% \label{fig:holdings_price}
% \end{figure}

\begin{figure}[h]
\centering
\vspace{-6px}
\includegraphics[width=\columnwidth]{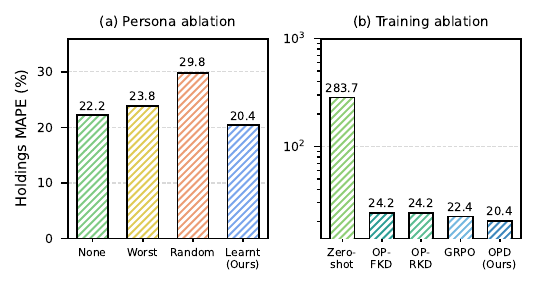}
\vspace{-25px}
\caption{Ablation studies for holdings (Lower is better).}
\label{fig:holdings_ablation}
\vspace{-4px}
\end{figure}

Figure~\ref{fig:holdings_ablation} repeats both ablations at the holdings level. We observe that the holdings MAPE is strongly correlated with the price MAPE in Tables~\ref{tab:ablation_role} and~\ref{tab:ablation_opd}, where \textsc{RoleCast} performs the best while the random personas and zero-shot agents performed the worst respectively.
This shows that the price gains are earned upstream at the demand-modeling stage, and improvements in simulated demand would transmit into improved price forecasts through the market-clearing identity.

\begin{figure}[h]
\centering
\vspace{-6px}
\includegraphics[width=0.99\columnwidth]{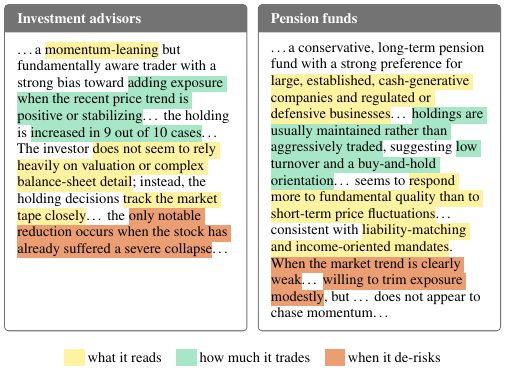}
\vspace{-8px}
\caption{Learnt personas from agent role-learning.}
\label{fig:learnt_personas}
\vspace{-7px}
\end{figure}

Figure \ref{fig:learnt_personas} shows examples of the learnt role personas from agent role-learning. Candidate personas were generated by prompting an LLM to infer the investor's investment style from randomly sampled input-output pairs, which are each optimized using GRPO, and the best persona is selected based on validation performance. The personas capture what information to focus on, how often to trade, and when to exit positions. We can observe clear differences across investor types: investment advisors exhibit a more aggressive, momentum-driven strategy, increasing exposure during positive price trends, while pension funds favor stable, cash-generating companies and conservative position adjustments. Thus, the role-learning step captures distinct investment behaviors for the agents.

% A complete list of the best learnt personas is found Appendix \ref{app:personas}, while case studies of their reasoning differences are in Appendix \ref{app:case_studies}.

\begin{figure}[h]
\vspace{-3px}
\centering
\begin{minipage}[c]{2.01in}
  \centering
  \includegraphics[width=\linewidth]{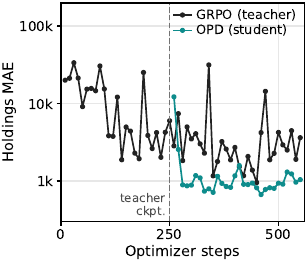}
\end{minipage}\hfill
\begin{minipage}[c]{1.20in}
  \vspace{-15px}
  \centering\small
  \setlength{\tabcolsep}{3pt}\renewcommand{\arraystretch}{0.91}
  \begin{tabular}{@{}lrr@{}}
    \toprule
    \textbf{SD} & GRPO & OPD \\
    \midrule
    \multicolumn{3}{@{}l}{\textit{Holdings}} \\
    \quad Value & 2,875 & 1,227 \\
    \quad Error & 1,421 &   339 \\
    \addlinespace[2pt]
    \multicolumn{3}{@{}l}{\textit{Price}} \\
    \quad Value &  23.6 &  17.7 \\
    \quad Error &  8.46 &  5.96 \\
    \bottomrule
  \end{tabular}
\end{minipage}
\vspace{-8px}
\caption{(\textit{Left}) MAE across training steps for the GRPO and OPD finetuned models on a matched-compute axis; (\textit{Right}) Standard deviation of the forecast values and forecast errors (MAE) for holdings and prices, at matched-compute levels.}
\vspace{-7px}
\label{fig:training}
\end{figure}

Figure \ref{fig:training} compares the training dynamics of GRPO and OPD under matched-compute levels. Here, the teacher model is first trained via GRPO for 250 steps. In parallel, we distill this policy onto a student model using OPD, while also continuing GRPO training on the same model separately, each for another 300 steps.

The left figure show that the OPD-trained model converges to a lower MAE with substantially smaller fluctuations, while right panel shows that OPD produces lower standard deviations in both its forecast values and errors (MAE), across holdings and prices. These results are consistent with the mode-seeking behavior of OPD (as described in Figure \ref{fig:opd-sharpening}), which concentrates the student on the teacher's preferred reasoning paths, whereas GRPO maintains a distribution of heuristically plausible paths, making it noisier.

\begin{figure}[h]
\centering
\vspace{-8px}
\hspace{-13px}
\includegraphics[width=1.04\columnwidth]{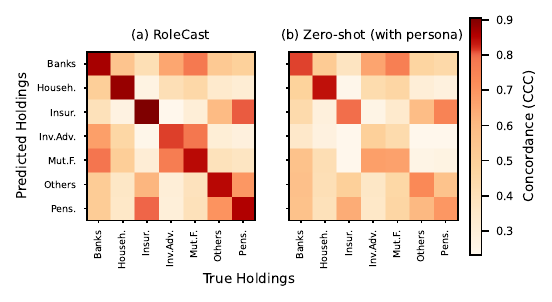}
\vspace{-20px}
\caption{Concordance between each agent's predicted holdings and the true realized holdings (Higher the better).}
\label{fig:agent_heatmaps}
\vspace{-4px}
\end{figure}

Figure~\ref{fig:agent_heatmaps} examines the concordance \cite{lawrence1989concordance} between forecasted and true holdings, which measures the correlation across both trends and scales. We find that each \textsc{RoleCast} agent matches its own institution's realized holdings best, \ie agents learn to match institution-specific behavior, rather than collapsing to an average investor. 

In contrast, the untrained model using only personas degrades sharply, showing that role descriptions alone mislead agents, and it is role-learning that converts them into faithful institution-level behavior. Since more accurate holdings lead to more accurate price forecasts, this behavioral fidelity can be seen as the source of the improved price gains themselves.

% \vspace{-2px}
\section{Conclusion}
In this work, we tackled stock price forecasting by modeling investor behavior instead of the price itself. To do this, we propose \textsc{RoleCast}, that produce price forecasts from simulated holdings through a closed-form market-clearing equation. To learn investor behavior, we utilize: agent role-learning, which searches for the persona that best reproduces the archetype's 13F disclosures, and on-policy distillation, which provides dense token-level feedback on the student's own reasoning traces. Experiments show that \textsc{RoleCast} outperforms state-of-the-art LLMs that forecast prices directly, while its agents faithfully reproduce institution-specific investment behavior. 

% Two future directions can be considered: extending the framework with finer-grained archetypes and higher-frequency disclosure data, and leveraging the released agents for counterfactual market simulation.
Two future directions can be considered for this work: Firstly, extending the framework with higher-frequency disclosure data, which can capture shorter-term changes in investor behavior; Secondly, leveraging the learnt investor agents for counterfactual market simulation, which could enable future studies on how different investor groups would respond to hypothetical market shocks.
% \vspace{-11px}

%%
%% The next two lines define the bibliography style to be used, and
%% the bibliography file.
\clearpage
\bibliographystyle{ACM-Reference-Format}
\bibliography{reference_long}
% \bibliography{reference}

%%
%% If your work has an appendix, this is the place to put it.
% \appendix
% \newpage
% \onecolumn
% \input{A_pricing}
% \input{a_significance}
% \input{b_data_construction}
% \input{c_contamination_control}
% \input{d_learnt_personas}
% \input{e_reasoning_case_studies}

\end{document}